\documentclass[runningheads]{llncs}

\usepackage[T1]{fontenc}

\usepackage{graphicx}
\usepackage{hyperref}
\usepackage{color}

\usepackage[misc]{ifsym}

\usepackage{bm}
\usepackage{pifont}
\usepackage{booktabs}
\usepackage{CJKutf8}
\usepackage{amsmath}
\usepackage{subcaption}
\usepackage{threeparttable}

\begin{document}

\title{Unsupervised Deep Cross-Language Entity Alignment}

\author{Chuanyu Jiang\inst{1\footnote[4]{}}     \and
        Yiming Qian\inst{2\footnote[4]{}}       \and
        Lijun Chen\inst{3}                      \and 
        Yang Gu\inst{4}                         \and
        Xia Xie\inst{1}(\Letter)}
        
\authorrunning{C. Jiang et al.}

\institute{School of Computer Science and Technology, Hainan University, China \\ \and 
Institute of High Performance Computing (IHPC), Agency for Science, Technology and Research (A*STAR), 1 Fusionopolis Way, \#16-16 Connexis, Singapore 138632, Republic of Singapore \\ \and
School of Cyberspace Security, Hainan University, China \\ \and
School of Information and Communication Engineering, Hainan University, China  \\
\email{\{cyhhyg, clara, guyangl, shelicy\}@hainanu.edu.cn} \\
\email{qiany@ihpc.a-star.edu.sg}}

\maketitle

\toctitle{Unsupervised Deep Cross-Language Entity Alignment}
\tocauthor{Chuanyu Jiang \and Yiming Qian \and Lijun Chen \and Yang Gu \and Xia Xie}

\renewcommand{\thefootnote}{\fnsymbol{footnote}}
\footnotetext[4]{First Author and Second Author contribute equally to this work.}

\begin{abstract}
Cross-lingual entity alignment is the task of finding the same semantic entities from different language knowledge graphs. In this paper, we propose a simple and novel unsupervised method for cross-language entity alignment. We utilize the deep learning multi-language encoder combined with a machine translator to encode knowledge graph text, which reduces the reliance on label data. Unlike traditional methods that only emphasize global or local alignment, our method simultaneously considers both alignment strategies. We first view the alignment task as a bipartite matching problem and then adopt the re-exchanging idea to accomplish alignment. Compared with the traditional bipartite matching algorithm that only gives one optimal solution, our algorithm generates ranked matching results which enabled many potentials downstream tasks. Additionally, our method can adapt two different types of optimization (minimal and maximal) in the bipartite matching process, which provides more flexibility. Our evaluation shows, we each scored 0.966, 0.990, and 0.996 $Hits@1$ rates on the $\mathrm{DBP15K}$ dataset in Chinese, Japanese, and French to English alignment tasks. We outperformed the state-of-the-art method in unsupervised and semi-supervised categories. Compared with the state-of-the-art supervised method, our method outperforms 2.6\% and 0.4\% in Ja-En and Fr-En alignment tasks while marginally lower by 0.2\% in the Zh-En alignment task.

\end{abstract}

\section{Introduction}
\begin{figure}[htbp!]
    \centering
    \includegraphics[width=0.95\textwidth]{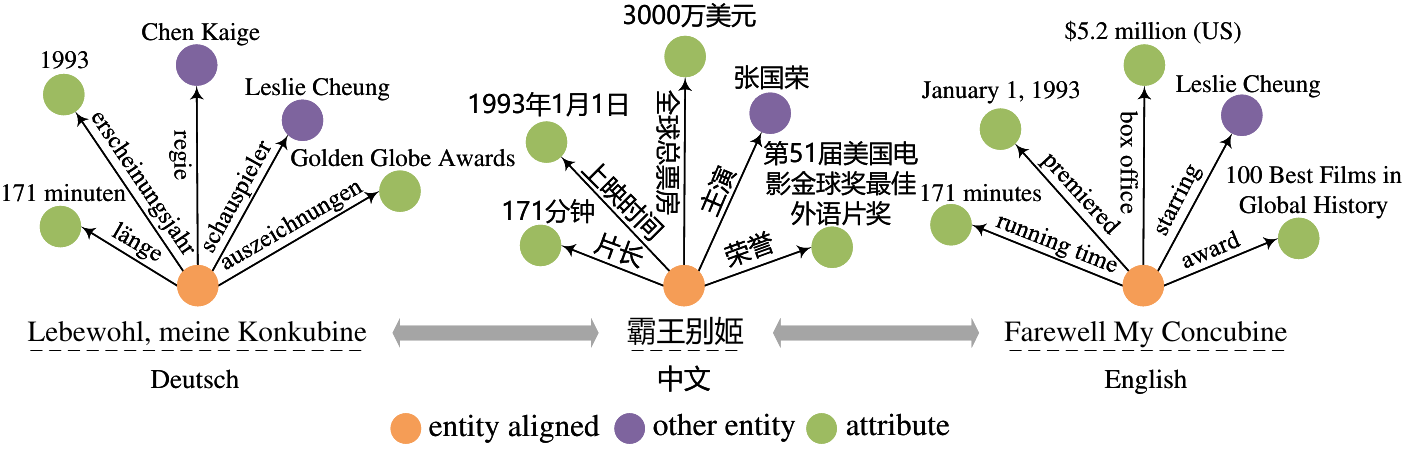}
    \caption{An illustration of between knowledge graphs cross-lingual entity alignment. The orange nodes are entities aligned across three knowledge graphs in three languages. The purple nodes are other entities that connect to the orange node, the green nodes are the attributes associated with the orange node.}
    \label{illustration}
\end{figure}

Knowledge graph (KG) was first proposed by Google in 2012~\cite{ikgts}, and it has been popularized in the fields of question answering, information retrieval, and recommendation system et al.~\cite{asakg}. Each knowledge graph contains structured information in the form of entities, relations, and semantic descriptions. Each entity is represented as a node in the graph, the relationship represents the relationship between connected entities, and the semantic description contains the attribute of each entity~\cite{askgraa}. The task of cross-lingual entity alignment is to find entities that are similar to the entities from a knowledge graph in another language. A schematic diagram of cross-lingual entity alignment between knowledge graphs is shown in Fig.~\ref{illustration}.

In recent years, deep learning methods are gaining popularity in cross-lingual entity alignment tasks. There are three major methods: supervised learning, semi-supervised learning, and unsupervised learning. The difference between those three methods is the percentage of labeled data available. The supervised methods require all data to be labeled and available in the training. When only a small subset of data is labeled, semi-supervised learning becomes the next optimal choice. It can utilize both labeled and unlabeled data in the training process. In many cases, having even a small amount of annotated data is a luxury where the unsupervised learning method becomes the go-to solution.

For existing unsupervised methods, we find these methods don't sufficiently utilize the extra information or do a relative complexity work to integrate extra information~\cite{eva,selfkg,ued,lightea,iclea}. Besides, by~\cite{seu,lightea} two methods we find that the global alignment can helps to improve accuracy but ignores the interaction with local information and lacks flexibility in the alignment process. Based on the above findings, we propose an unsupervised method called \textbf{U}nsupervised \textbf{D}eep \textbf{C}ross-Language \textbf{E}ntity \textbf{A}lignment (UDCEA). Our method conducts the embedding process with combinations of simple machine translation and pre-training language model encoder, which avoids the use of complex technologies like Graph Attention Networks (GAT)~\cite{gan}, graph convolutional network (GCN)~\cite{gcn}, et al. In the alignment process, our method makes the global and local information interact. Our alignment strategy not only brings accuracy increases but also improves flexibility in the alignment process. Our method is divided into two modules: feature embedding and alignment.

In the feature embedding module, we are not limited to using the entity name information but further utilize the multi-view information (entity name, structure, and attribute). On the multi-view information, we apply the machine translation (Google translator) engine to translate non-English text into English, which provides a weak alignment between entities across two languages. This practice is quite popular in the unsupervised entity alignment methods~\cite{ued,seu,rdgcn,raga}. Then we apply a pre-trained multi-language encoder~\cite{msem} to obtain embedding for multi-view information. From our experiments, we found the encoder model trained in a multi-language fashion scored higher alignment accuracy than the monolingual encoder when the two languages have a distinct difference. 

In the alignment module, we are not only adopting the global alignment to conduct alignment and further combining it with the local alignment. The first step is to construct the adjacency matrix for global alignment. The adjacency matrix is built by a simple but efficient method, which is fusing the multi-view information into a unified matrix. Usually, the preliminary adjacency matrix does not have the directional information and so we transposed the adjacency matrix to add the directional information. Next, we utilize the optimization algorithm to handle the adjacency matrix containing directional information, which attains the global information. Finally, we allow the global information interaction with local information (specific operations in~\ref{alignment module}) and get the final ranking alignment results. Our alignment strategy can be deployed to improve existing approaches without heavy modifications. The detailed evaluation is in section~\ref{sec: additional analysis}. Our evaluation shows, we improve 1.0\% to 5.6\% for SEU~\cite{seu}, 15.3\% to 16.5\% for RREA~\cite{rrea}, and 9.9\% to 18.3\% for HGCN-JE~\cite{hgcn-je-jr} on three language alignment tasks. Furthermore, our method can adapt to different data sizes, which the detailed experiments result is shown in Fig.~\ref{fig: additional analysis}(\subref{fig: facing data changes}).

We evaluated our overall alignment method on the $\mathrm{DBP15K}$ dataset~\cite{DBP15K} which results achieving $0.966$, $0.990$, and $0.996$ accuracy each in Zh-En, Ja-En, and Fr-En datasets. We exceed the state-of-the-art unsupervised and semi-supervised methods in alignment accuracy. Even compared with the state-of-the-art supervised method, our method surpasses them by 2.6\% and 0.4\% in Ja-En and Fr-En alignment while only marginally lower by 0.2\% in the Zh-En alignment. In conclusion, our method contributions are summarized as follows:

\begin{itemize}
    \item[$\bullet$] We apply a simple but efficient method to fuse the multi-view information into one unified adjacency matrix.
    \item[$\bullet$] We propose a novel method that extracts the ranked matching results from state-of-the-art bipartite matching algorithms.
    \item[$\bullet$] We conduct throughout experiments to study the impact of different configurations of machine translation and encoders on the entity alignment task.
    \item[$\bullet$] We show our alignment module method is flexible and can be used as an add-on to improve existing entity alignment methods.
\end{itemize}

\section{Related Work}
The early entity alignment methods~\cite{tfidf,asmov,logmap} were mainly based on measuring the text similarity between entity pairs augmented by relation information and probability models. These early methods tend to ignore the hide semantic information while suffering scalability issues. As the size of available data increases, so does the degree of knowledge graph complexity. The urge of finding a method to handle large amounts of data leads to wide adaptation of data-driven learning approaches. Based on the amount of annotated available data, we can categorize the learning based approaches into supervised, semi-supervised, and unsupervised three types.

\subsection{Supervised Entity Alignment}
The supervised entity alignment method requires access to fully annotated data for training. The early works deploy translation-based methods such as TransE~\cite{transe}, TransH~\cite{transh}, TransR~\cite{transr}, and TransD~\cite{ji-etal-2015-knowledge} to obtain the feature embedding. TransE assumes each entity $\boldsymbol{h}$ on the graph is a linear combination of its adjacent entity $\boldsymbol{t}$ and relation $\boldsymbol{r}$ which can be formulated as $\boldsymbol{h}+\boldsymbol{r}\approx \boldsymbol{t}$. This linear combination approach suffers huge errors when the entities have a 1-to-N or N-to-N relation with other entities. TransH reduced this problem by projecting the entity vector onto a hyperplane which preserved the topology information between vectors. Both TransE and TransH assume the vector of entities and relations in the same space. However, multiple attributes may associate with the same entity so this assumption may not always hold. TransR removed this assumption by projecting the entity and relation vector into different spaces. TransD took a more in-depth division for the projection between entity and relation.

The linear combination between vectors is fast but as dataset size grows, the complexity between entities grows exponentially. To handle this growing complexity, GCN based methods~\cite{gcn-align,cg-mualign} start to gain popularity. The drawback of the learning-based entity alignment method is the requirement of an enormous amount of labeled data to properly train a model. One way to reduce this drawback is through semi-supervised learning.

\subsection{Semi-supervised Entity Alignment}
The semi-supervised method solved the label requirement only by utilizing a small number of alignment labels as seed. Then this seed propagates over the knowledge graph to align the rest of the entities. To address the labeled data relative lack problem, semi-supervised alignment methods use the iterative strategy to generate new training data~\cite{mraea,xin2022ensemble,naea}. With the iterative process starting, more and more pseudo-labeled data will be added to the training sets until reaching the stop criteria. The semi-supervised method relaxed the reliance on the amount of labeled data and this data requirement can be further reduced to zero by unsupervised learning.

\subsection{Unsupervised Entity Alignment}
Unsupervised methods apply external resources to eliminate the need of requiring labeled data in the entity alignment. One popular external resource is machine translation engine~\cite{ued,seu,lightea,rdgcn,raga} such as Google translator. Non-English entities are translated into English and then encoded into text embedding. The encoding methods can be statistical  (e.g. N-gram~\cite{n-gramsd}, and GloVe~\cite{glove}), or deep learning  (e.g. BERT~\cite{bert}, and RoBERTa~\cite{roberta}). Later,~\cite{selfkg} suggests these two-step machine translation and encoding process can be replaced by a single pre-trained multi-lingual model~\cite{labse}. The external resources can be used to obtain the semantic embedding and further generate the similarity matrix. Finally, the similarity matrix is processed by optimization algorithms such as Hungarian algorithm~\cite{seu}, deferred acceptance algorithm~\cite{raga}, or Sinkhorn algorithm~\cite{seu,lightea} to conduct alignment.

\section{Proposed Method}
\begin{figure*}[htbp!]
    \centering
    \includegraphics[width=0.95\textwidth]{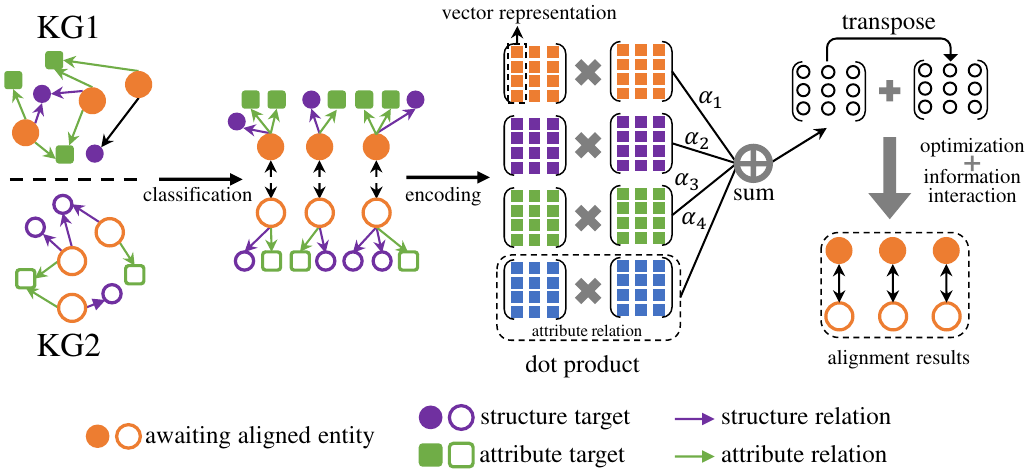}
    \caption{The overall flow of our method.}
    \label{fig: overall flow}
\end{figure*}

\renewcommand{\thefootnote}{\arabic{footnote}}
Our method\footnote[1]{Our source code is available in~\href{https://github.com/chuanyus/UDCEA}{https://github.com/chuanyus/UDCEA.}} contains two modules: feature embedding and alignment. The feature embedding module utilizes the machine translator and pre-trained deep multi-language learning encoder to generate feature embedding for each aligned entity. The alignment module first generates the similarity matrix based on feature embedding and then applies the optimization-based method to align two knowledge graphs. The overall flow of our method is shown in Fig.~\ref{fig: overall flow}.

\subsection{Base Symbol Definition}
Given two knowledge graphs in different languages, named $G_{1}$ and $G_{2}$. $E_{1} =\left \{ e_{(1,1)},e_{(1,2)},...,e_{(1,m_{1})} \right \}$ and $E_{2} =\left \{ e_{(2,1)},e_{(2,2)},...,e_{(2,m_{2})} \right \}$ denote a set of entity in $G_{1}$ and $G_{2}$ respectively, which $m_{1}$ and $m_{2}$ is the index. Abstractly, we define $E$ to represent entity name information. The goal is to find all entity pairs that share the same semantic information in $E_{1}$ and $E_{2}$. An entity of knowledge graph usually has two types of information which are the structure and attribute information. The structure information represents the relation between entities while attribute information usually is to entity description. The attribute information represents the properties that associate with each entity. We further divide the structure and attribute information into target and relation information. For example in Fig.~\ref{fig: overall flow}, the purple nodes, green nodes, purple arrowheads, and green arrowheads are the structure target (ST), attribute target (AT), structure relation (SR), and attribute relation (AR) respectively. Here, we denote $ST_{1}$, $SR_{1}$, $AT_{1}$, and $AR_{1}$ to represent the information from $G_{1}$. Similarly, we denote $ST_{2}$, $SR_{2}$, $AT_{2}$, and $AR_{2}$ to represent the information from $G_{2}$.

\subsection{Feature Embedding Module}
The feature embedding module consists of two steps which are language translation and text encoding. In this process, we choose the $E$, $ST$, $AT$, and $AR$ for embedding and give up the $SR$ feature. Because the $SR$ information generally exists in reality so lacks identity property. In the language translation step, we convert all non-English text into English using Google translator $f_{t}$. This translation process can be formulated as: 
\begin{equation}
    (E_{i}^{'},ST_{i}^{'},AT_{i}^{'},AR_{i}^{'})=f_{t}\left( (E_{i},ST_{i},AT_{i},AR_{i}) \right), \quad i=1,2.
\end{equation}
Where $E_{i}^{'}$, $ST_{i}^{'}$, $AT_{i}^{'}$, and $AR_{i}^{'}$ is the translated text. 

Next, we utilize the multi-language Sentence-BERT~\cite{msem} encoder $f_{e}$ to encode these texts. The encoder maps the text to a 768-dimensional dense vector that includes abundant semantic information. This encoding process is formulated as:
\begin{equation}
    (V_{(i,E)},V_{(i,ST)},V_{(i,AT)},V_{(i,AR)})=f_{e}(E_{i}^{'},ST_{i}^{'},AT_{i}^{'},AR_{i}^{'}),\quad i=1,2.
\end{equation}
Where $V_{(i,E)}$, $V_{(i,ST)}$, $V_{(i,AT)}$, and $V_{(i,AR)}$ is the feature embedding for $E$, $ST$, $AT$ and $AR$ respectively.

\subsection{Alignment Module}
\label{alignment module}
Taking the embedding generated by the feature embedding module, we get the set of vectors $V_{(i,E)}$, $V_{(i,ST)}$, $V_{(i,AT)}$, and $V_{(i,AR)}$ for $E$, $ST$, $AT$, and $AR$ respectively:
\begin{equation}
    \left\{\begin{aligned}
    V_{(i, E)}  &=  (\boldsymbol{{e}_{(i,1)}} , \boldsymbol{{e}_{(i,2)}} , ... , \boldsymbol{{e}_{(i,m_{i})}}) ,  \\
    V_{(i, ST)} &=  (\boldsymbol{{st}_{(i,1)}}, \boldsymbol{{st}_{(i,2)}}, ... , \boldsymbol{{st}_{(i,m_{i})}}),  \\ 
    V_{(i, AT)} &=  (\boldsymbol{{at}_{(i,1)}}, \boldsymbol{{at}_{(i,2)}}, ... , \boldsymbol{{at}_{(i,m_{i})}}),  \\ 
    V_{(i, AR)} &=  (\boldsymbol{{ar}_{(i,1)}}, \boldsymbol{{ar}_{(i,2)}}, ... , \boldsymbol{{ar}_{(i,m_{i})}}),
    \end{aligned} \right. \quad i=1,2.
\end{equation}

Next, the pairwise similarity matrices $S_{E}$, $S_{ST}$, $S_{AT}$ and $S_{AR}$ for $E$, $ST$, $AT$, and $AR$ are constructed by calculating the dot product between normalized vectors:
\begin{equation}
    \left\{\begin{aligned}
    S_{E}  &= (\boldsymbol{e_{(1,1)}}, \boldsymbol{e_{(1,2)}}, ... , \boldsymbol{e_{(1,m_{1})}})^{T}    (\boldsymbol{e_{(2,1)}}, \boldsymbol{e_{(2,2)}}, ... , \boldsymbol{e_{(2,m_{2})}}),                \\ 
    S_{ST} &= (\boldsymbol{{st}_{(1,1)}}, \boldsymbol{{st}_{(1,2)}}, ... , \boldsymbol{{st}_{(1,m_{1}}})^{T} ( \boldsymbol{{st}_{(2,1)}}, \boldsymbol{{st}_{(2,2)}}, ..., \boldsymbol{{st}_{(2,m_{2})}}),  \\
    S_{AT} &= (\boldsymbol{{at}_{(1,1)}}, \boldsymbol{{at}_{(1,2)}}, ... , \boldsymbol{{at}_{(1,m_{1})}})^{T} ( \boldsymbol{{at}_{(2,1)}}, \boldsymbol{{at}_{(2,2)}}, ..., \boldsymbol{{at}_{(2,m_{2})}}),  \\
    S_{AR} &= (\boldsymbol{{ar}_{(1,1)}}, \boldsymbol{{ar}_{(1,2)}}, ... , \boldsymbol{{ar}_{(1,m_{1})}})^{T} ( \boldsymbol{{ar}_{(2,1)}}, \boldsymbol{{ar}_{(2,2)}}, ..., \boldsymbol{{ar}_{(2,m_{2})}}).
    \end{aligned} \right.
\end{equation}

The weighted summation of the similarity matrices forms a new matrix $S$ which can be used as an adjacency matrix for the later alignment task. We denote the weight parameter $\alpha_{E}$, $\alpha_{ST}$, $\alpha_{AT}$, and $\alpha_{AR}$ for $S_{E}$, $S_{ST}$, $S_{AT}$ and $S_{AR}$ respectively. The summation step be expressed as:
\begin{equation}
    S = (S_{E},S_{ST},S_{AT},S_{AR})(\alpha_{E},\alpha_{ST},\alpha_{AT},\alpha_{AR})^{T}.
\end{equation}

Where $S$ is the new matrix with dimension of $m_{1}\times m_{2}$. For the case of $S$ is not a square matrix, we add padding to the shorter dimension to convert it into a square matrix. The value in the padding depends on the optimization strategy (minimal or maximal optimization), where positive and negative infinity are used for minimal and maximal optimization respectively. In this way, we can minimize the impact of the padding and get the matrix $\bar S$ that dimension is the $n\times n$ ($n$ is the maximum value in $m_{1}$ and $m_{2}$). The~\cite{raga} indicates that directional information can improve alignment accuracy but we notice that the softmax operation is an unnecessary step (specific in~\ref{alignment module analysis}). So the transpose of $\bar S$ is directly added to the $\bar S$ and forms a new adjacency matrix $A = \bar S + \bar S^{T}$ that contains directional information.

\begin{figure}[htbp!]
    \centering
    \includegraphics[width=0.92\textwidth]{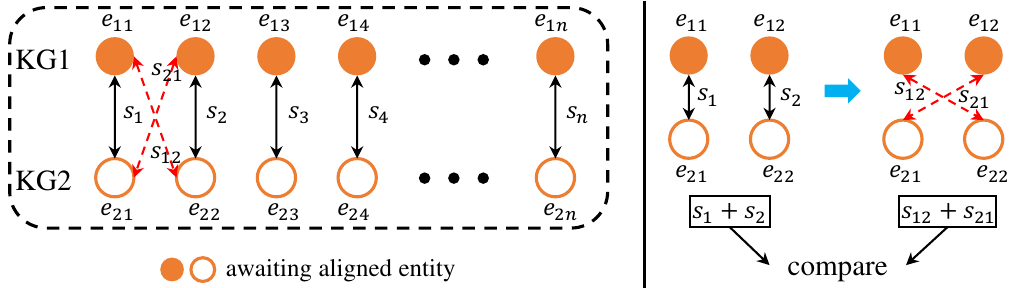}
    \caption{A single entity alignment example. The orange nodes are the awaiting aligned entities that use $e_{11},e_{12},...$ to represent. The $s_{1},s_{2},...$ are the similarity degree values between aligned entities.}
    \label{alignment example}
\end{figure}

Unlike previous methods that only utilize global or local information in alignment, we employ both global and local information to interact. Our method first utilizes an optimization algorithm to get the preliminary alignment results and based this to further align the single entity. Before giving formulas of the final weighting matrix generating, we show an example in Fig.~\ref{alignment example}.

In Fig.~\ref{alignment example}, the $(e_{11},e_{21})$ and $(e_{12},e_{22})$ are the alignment entity pairs that total similarity degree is $s_{1}+s_{2}$. The $e_{11}$ and $e_{12}$ aligns with $e_{22}$ and $e_{21}$ respectively and receive a new similarity value $s_{12}+s_{21}$. We calculate the value of $(s_{12}+s_{21})-(s_{1}+s_{2})$ as the loss of this exchanging operation. The value of loss is greater which means this exchanging more effectivity. We repeat the above exchanging process until exchange all aligned entities in the preliminary alignment results. This step gets a set of losses which includes all the exchanging losses from each step. Then, we sort the loss set and get the $e_{11}$ alignment sequence.

Based on the above example, the weighting matrix generates formulas that are the description for all single entity exchanging process calculations. Firstly, the similarity degree of preliminary alignment results needs in the diagonal of matrix $A$. So we use the preliminary alignment results to re-arrange matrix A and get matrix $\bar A$. Next, we construct the auxiliary matrix $U$ as:
\begin{equation}
    U=\begin{bmatrix}
        s_{1}  & s_{1}  & \dots  & s_{1}    \\
        s_{2}  & s_{2}  & \dots  & s_{2}    \\
        \vdots & \vdots & \dots  & \vdots   \\
        s_{n}  & s_{n}  & \dots  & s_{n}
    \end{bmatrix}
\end{equation}
Where the $\left \{s_{1},s_{2},...,s_{n} \right \}$ is the diagonal elements of $\bar A$. 

Finally, we attain the weighting matrix $A_{unsorted}$ by the following formula:
\begin{equation}
    \begin{aligned}
        A_{unsorted} &= (\bar A+\bar A^{T})-(U+U^{T}) \\
                     &= (\bar A-U)+(\bar A-U)^{T}.
    \end{aligned}
\end{equation}

Then we sort the elements of each row in $A_{unsorted}$ and obtain a sorted matrix $A_{sorted}$ which indicates the ranked order for the top matching results.
\section{Experiments}
\subsection{Cross-lingual Dataset}
We pick DBP15K~\cite{DBP15K} dataset to evaluate our method. This dataset contains knowledge graphs from four languages which are English (En), Chinese (Zh), and Japanese (Ja), and French (Fr). Three language pairs are evaluated which are Zh-En, Ja-En, and Fr-En. Each language pair contains about 15,000 entity pairs.

\subsection{Comparative}
To benchmark the efficiency of our method, we choose 20 state-of-art methods as a comparison. Three categories of methods are evaluated which are supervised, semi-supervised, and unsupervised learning. The supervised learning which utilizes a full training set to conduct alignment, are MuGNN~\cite{mugnn}, AliNet~\cite{alinet}, HGCN-JE~\cite{hgcn-je-jr}, RpAlign~\cite{rpalign}, SSEA~\cite{ssea}, RNM~\cite{rnm}, FuzzyEA~\cite{fuzzyea}, HMAN~\cite{hman}, EPEA~\cite{epea}, and BERT-INT~\cite{bert-int}. The semi-Supervised learning method, which only utilized a subset of the training set, are KECG~\cite{kegg}, NAEA~\cite{naea}, MRAEA~\cite{mraea}, RREA~\cite{rrea}. The unsupervised learning, which doesn't need a training set, are EVA~\cite{eva}, SelfKG~\cite{selfkg}, UED~\cite{ued}, SEU~\cite{seu}, ICLEA~\cite{iclea}, and LightEA~\cite{lightea}.

\begin{table*}[ht!]
\setlength{\tabcolsep}{0.5pt}
\caption{Comparison of entity alignment results on DBP15K dataset.}
\label{tab: full_results}
    \begin{threeparttable}
    \scalebox{0.96}{
    \begin{tabular}{c|ccc|ccc|ccc}
        \toprule[1.1pt]
        & \multicolumn{3}{c}{$\mathrm{DBP15K_{Zh-En}}$}  & \multicolumn{3}{|c|}{$\mathrm{DBP15K_{Ja-En}}$}      & \multicolumn{3}{c}{$\mathrm{DBP15K_{Fr-En}}$}                                            \\ \toprule[1.1pt]
        Method     & $Hits@1$        & $Hits@10$         & MRR            & $Hits@1$         & $Hits@10$      & MRR             & $Hits@1$         & $Hits@10$      & MRR                                \\ \hline
        \multicolumn{10}{c}{Supervised}         \rule{0pt}{11pt}                                                                                                                                         \\ \hline
        MuGNN~\cite{mugnn}              & 0.494          & 0.844          & 0.611            & 0.501          & 0.857           & 0.621            & 0.495          & 0.870          & 0.621             \\
        AliNet~\cite{alinet}            & 0.539          & 0.826          & 0.628            & 0.549          & 0.831           & 0.645            & 0.552          & 0.852          & 0.657             \\
        HGCN-JE~\cite{hgcn-je-jr}       & 0.720          & 0.857          & -                & 0.766          & 0.897           & -                & 0.892          & 0.961          & -                 \\
        RpAlign~\cite{rpalign}          & 0.748          & 0.889          & 0.794            & 0.730          & 0.890           & 0.782            & 0.752          & 0.900          & 0.801             \\
        SSEA~\cite{ssea}                & 0.793          & 0.899          & -                & 0.830          & 0.930           & -                & 0.918          & 0.972          & -                 \\
        RNM~\cite{rnm}                  & 0.840          & 0.919          & 0.870            & 0.872          & 0.944           & 0.899            & 0.938          & 0.981          & 0.954             \\
        FuzzyEA~\cite{fuzzyea}          & 0.863          & 0.984          & 0.909            & 0.898          & 0.985           & 0.933            & 0.977          & $\bm{0.998}$   & 0.986             \\
        HMAN~\cite{hman}                & 0.871          & 0.987          & -                & 0.935          & $\bm{0.994}$    & -                & 0.973          & $\bm{0.998}$   & -                 \\
        EPEA~\cite{epea}                & 0.885          & 0.953          & 0.911            & 0.924          & 0.969           & 0.942            & 0.955          & 0.986          & 0.967             \\
        BERT-INT~\cite{bert-int}        & $\bm{0.968}$ & $\bm{0.990}$     & $\bm{0.977}$     & $\bm{0.964}$   & 0.991           & $\bm{0.975}$     & $\bm{0.992}$   & $\bm{0.998}$   & $\bm{0.995}$      \\
        \hline
        \multicolumn{10}{c}{Semi-Supervised}    \rule{0pt}{11pt}                                                                                                                                         \\ \hline
        KECG~\cite{kegg}                & 0.478          & 0.835          & 0.598            & 0.490          & 0.844           & 0.610            & 0.486          & 0.851          & 0.610             \\
        NAEA~\cite{naea}                & 0.650          & 0.867          & 0.720            & 0.641          & 0.873           & 0.718            & 0.673          & 0.894          & 0.752             \\
        MRAEA~\cite{mraea}              & 0.757          & 0.930          & 0.827            & 0.758          & 0.934           & 0.826            & 0.780          & 0.948          & 0.849             \\
        RREA~\cite{rrea}                & $\bm{0.801}$   & $\bm{0.948}$   & $\bm{0.857}$     & $\bm{0.802}$   & $\bm{0.952}$    & $\bm{0.858}$     & $\bm{0.827}$   & $\bm{0.966}$   & $\bm{0.881}$      \\ \hline
        \multicolumn{10}{c}{Unsupervised}   \rule{0pt}{11pt}                                                                                                                                             \\ \hline
        EVA~\cite{eva}                  & 0.761          & 0.907          & 0.814            & 0.762          & 0.913           & 0.817            & 0.793          & 0.942          & 0.847             \\
        SelfKG~\cite{selfkg}            & 0.745          & 0.866          & -                & 0.816          & 0.913           & -                & 0.959          & 0.992          & -                 \\
        UED~\cite{ued}                  & 0.826          & 0.943          & 0.870            & 0.863          & 0.960           & 0.900            & 0.938          & 0.987          & 0.957             \\
        SEU~\cite{seu}                  & 0.900          & 0.965          & 0.924            & 0.956          & 0.991           & 0.969            & 0.988          & $\bm{0.999}$   & 0.992             \\
        ICLEA~\cite{iclea}              & 0.921          & 0.981          & -                & 0.955          & 0.988           & -                & 0.992          & $\bm{0.999}$   & -                 \\
        LightEA-L~\cite{lightea}        & 0.952          & 0.984          & 0.964            & 0.981          & $\bm{0.997}$    & 0.987            & 0.995          & 0.998          & 0.996             \\ 
        \textbf{UDCEA (ours)}           & $\bm{0.966}$   & $\bm{0.989}$   & $\bm{0.975}$     & $\bm{0.990}$   & $\bm{0.997}$    & $\bm{0.993}$     & $\bm{0.996}$   & $\bm{0.999}$   & $\bm{0.997}$      \\ \toprule[1.1pt]
    \end{tabular}}
    \scriptsize{*The comparison results are extracted from the corresponding paper.}
    \end{threeparttable}
\end{table*}

\subsection{Evaluation Indicate and Experiment Setting}
We use $Hits@n$ and MRR (Mean Reciprocal Rank) to evaluate our approach. $Hits@n$ represents the proportion of truly aligned entity ranking greater or equal to $n$ and MRR is the sum of the reciprocal ranks of all aligned entities divided by the total number of aligned entities. In the embedding generation module, we use Google translator and multi-language encoder~\cite{msem} as our translator and encoder respectively. In the entity alignment module, the value of $\alpha_{E}$, $\alpha_{ST}$, $\alpha_{AT}$, and $\alpha_{AR}$ are set to 1.00, 0.75, 0.75, and 0.15 respectively. We choose the Jonker-Volgenant algorithm~\cite{lapjv} to conduct a bipartite matching process.

The weight of the four parameters is according to their significance. Naturally, the E most represents the aligned entity, the next is ST and  AT, and the final is AR. Based on this idea, we set the weight for the four parameters respectively. In fact, the weight doesn't have a large impact on the alignment result (specific experiments in~\ref{alignment module analysis}). For the optimization algorithm, we choose the Jonker-Volgenant algorithm because it has a relatively fast speed and stability. The time complexity of the Jonker-Volgenant algorithm is $O(n^{3})$. If a faster speed is needed, Sinkhorn algorithm~\cite{seu,lightea,sinkhorn} is the next best option.

\subsection{Experimental Results}
Our results are shown in Table~\ref{tab: full_results}. Experiment results show our method exceeds all baselines in $Hits@1$, $Hits@10$, and MRR and achieves state-of-the-art results. Specifically, compared with unsupervised methods our method outperformed the next highest algorithm by 1.4\%, 0.9\%, and 0.1\% in the $Hits@1$ accuracy. Even for relatively difficult $\mathrm{DBP15K_{Zh-En}}$ dataset, we also achieve high accuracy of 0.966. To a certain extent, it shows our method's ability to effectively handle difficult cross-language entity alignment tasks. Moreover, we still obtain an advantage compared with state-of-art semi-supervised and supervised methods. Compared with semi-supervised methods, our method exceeds all methods and keeps an average improvement of above 15\% accuracy. Compared with the state-of-the-art supervised method BERT-INT~\cite{bert-int}, we still surpass it by 2.6\% and 0.4\% for $\mathrm{DBP15K_{Ja-En}}$ and $\mathrm{DBP15K_{Fr-En}}$ datasets and only marginally lower by 0.2\% in $\mathrm{DBP15K_{Zh-En}}$ dataset. These experimental results show the superior performance of our method.

\section{Ablation Study}
\label{ablation study}
In this section, we present a detailed analysis of our method. The total description is as follows:
\begin{itemize}
    \item[$\bullet$] Analysis the influence of translator and encoder on results (\ref{translator and encoder analysis}).
    \item[$\bullet$] Analysis the influence of E, ST, AT, and AR on results (\ref{alignment module analysis}).
    \item[$\bullet$] Analysis the influence of bi-directional information on results (\ref{alignment module analysis}).
    \item[$\bullet$] Analysis the influence of minimal or maximal optimization on results (\ref{alignment module analysis}).
    \item[$\bullet$] Evaluation the influence of our alignment method on existing methods (\ref{sec: additional analysis}).
    \item[$\bullet$] Study the adaptability of our method on data sizes (\ref{sec: additional analysis}).
\end{itemize}

\subsection{Translator and Encoder Analysis}
\label{translator and encoder analysis}
We analyze the influence of Google translator and multi-language encoder~\cite{msem} on the entity alignment task. The experiment results are shown in Table~\ref{tab: 2}. For translators, we give up using Google translator and directly utilize the multi-language encoder to encode. The experiment results show that Google translator improves 3.5\% and 5.4\% accuracy on the Zh-En and Ja-En datasets respectively while decreasing 0.2\% accuracy on the Fr-En dataset. It indicates that the translator helps in cross-language entity alignment when the two languages have a distinct difference while the translator has a minimal influence when the two languages are similar. For the encoder, we choose N-Gram~\cite{n-gramsd}, GloVe~\cite{glove}, and a monolingual sentence-bert~\cite{sentence-bert} model as the comparison. The experiment results show that the multi-language encoder has an absolute advantage compared with the N-Gram and GloVe encoders, which gives the accuracy a large boost. Compared with the monolingual encoder, the multi-language encoder still improves 3\%, 1.1\%, and 1.2\% accuracy on three datasets. In conclusion, the Google translator and multi-language encoder both have a positive influence on entity alignment mission, which combined can achieve higher accuracy.

\begin{table}[!htbp]
\small
\centering
\renewcommand\arraystretch{1.15}
\caption{The experiment on translator and encoder. For parameter setting, the n is 2 in the N-Gram model and the dimension is 300 in the GloVe model. The symbol description: MS-B denotes multi-language Sentence-BERT~\cite{msem}, S-B denotes monolingual Sentence-Bert~\cite{sentence-bert}.}
\begin{tabular}{cccccccc} \toprule[1.1pt]
                            &                 & \multicolumn{2}{c}{Zh-En}                 & \multicolumn{2}{c}{Ja-En}                 & \multicolumn{2}{c}{Fr-En}                  \\ \toprule[1.1pt]
    Encoder Model           & Translation     & $Hits@1$          & $Hits@10$             & $Hits@1$          & $Hits@10$             & $Hits@1$              & $Hits@10$          \\ \hline
    N-Gram                  & \ding{51}       & 0.352             & 0.523                 & 0.425             & 0.602                 & 0.550                 & 0.670              \\ \hline
    GloVe                   & \ding{51}       & 0.718             & 0.841                 & 0.791             & 0.896                 & 0.840                 & 0.915              \\ \hline
    S-B                     & \ding{51}       & 0.936             & 0.978                 & 0.979             & 0.994                 & 0.984                 & 0.994              \\ \hline
    MS-B                    & \ding{51}       & $\bm{0.966}$      & $\bm{0.989}$          & $\bm{0.990}$      & $\bm{0.997}$          & 0.996                 & 0.999              \\ \hline
    MS-B                    & \ding{53}       & 0.931             & 0.987                 & 0.936             & 0.978                 & $\bm{0.998}$          & $\bm{1.000}$       \\ \bottomrule[1.1pt]
\end{tabular}
\label{tab: 2}
\end{table}

\subsection{Alignment module Analysis}
\label{alignment module analysis}
\begin{figure}
\centering
\begin{subfigure}[t]{0.495\linewidth}
    \centering
    \includegraphics[width=\linewidth]{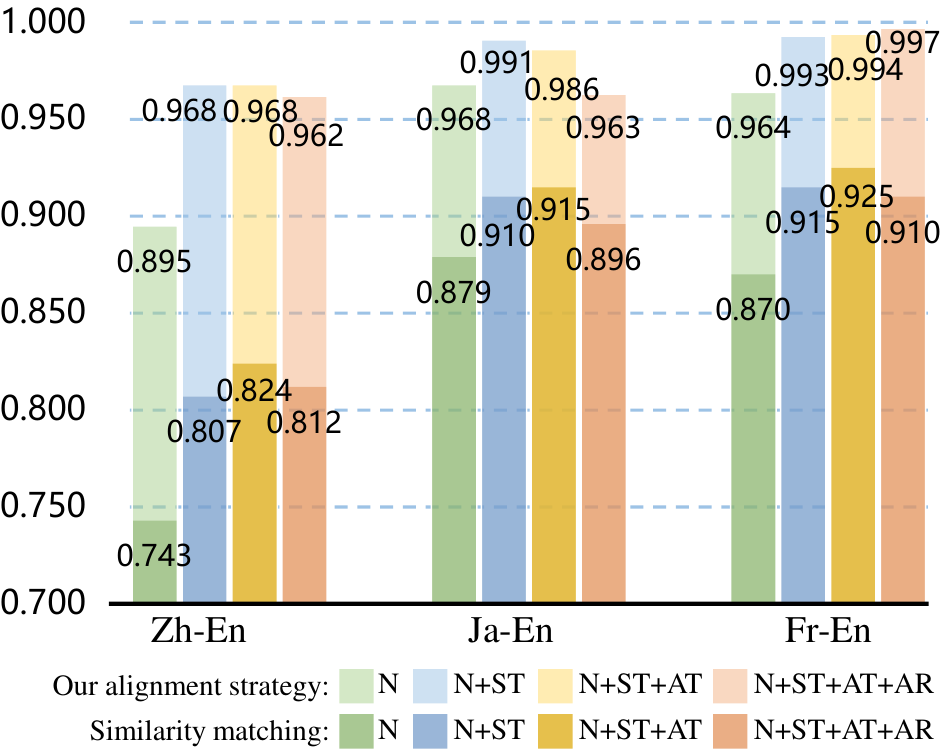}
    \subcaption{The influence of multi-view information and alignment strategy for accuracy.}  
    \label{fig: multi-view and alignment}
\end{subfigure}
\begin{subfigure}[t]{0.495\linewidth}
    \centering
    \includegraphics[width=0.955\linewidth]{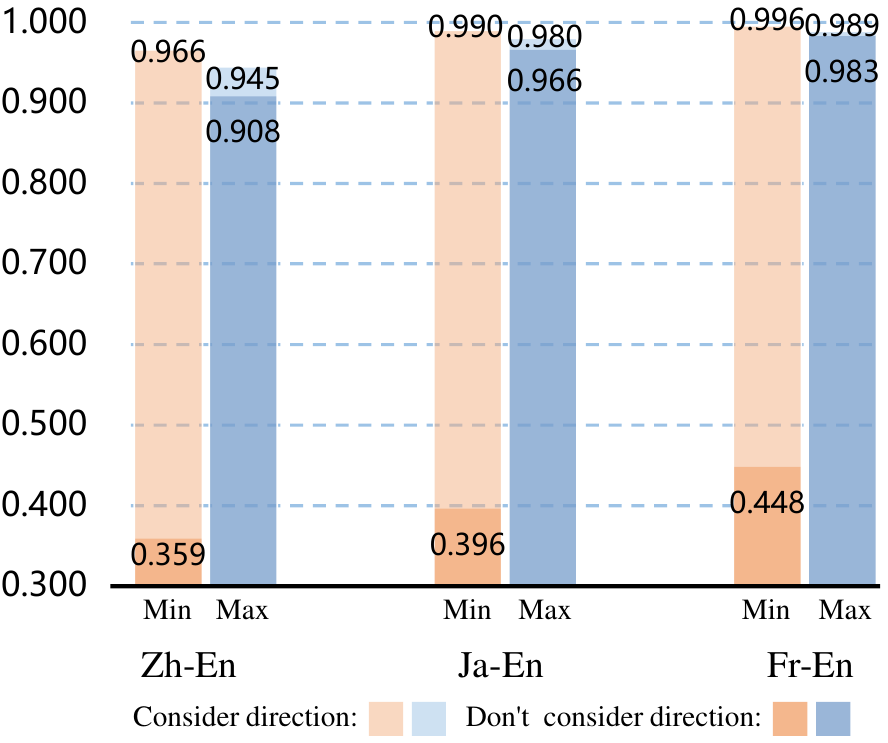}
    \subcaption{The influence of minimal and maximal optimization for accuracy.}
    \label{fig: minimal or maximal optimization}
\end{subfigure}
\caption{The experiment results of alignment module analysis.}
\label{fig: alignment module analysis}
\end{figure}

We analyze the influence of multi-view information and alignment strategy for accuracy. To further prove the advantage of our alignment strategy, we add the similarity matching alignment method (directly sorted by similarity degree value) as a comparison in all experiments of this part. The experiment results are shown in Fig.~\ref{fig: alignment module analysis}. In Fig.~\ref{fig: alignment module analysis}(\subref{fig: multi-view and alignment}), the weight of N, ST, AT, and AR is set to 1.00. 

For multi-view information. From Fig.~\ref{fig: alignment module analysis}(\subref{fig: multi-view and alignment}), we can see that the alignment accuracy gradually increases as the multi-view information is added. The experiment results indicate that the multi-view information is beneficial for improving accuracy. Meanwhile, notice that after adding the ST information that the alignment results have already achieved outstanding results. It indicates our method has enough robustness that can adapt to the multi-view information part lack of circumstance. What's more, when we do not consider the weighting of multi-view information (all the weights set to 1.00), we can find that the accuracy may decrease as the multi-view information is added, which indicates that weighting adjustment is necessary. For alignment strategy. In Fig.~\ref{fig: alignment module analysis}(\subref{fig: multi-view and alignment}), it shows that our alignment strategy achieves higher accuracy than the similarity matching method in all experiments. 

The evaluations of the influence of directional information, and minimal, and maximal optimization for alignment results are displayed in Fig.~\ref{fig: alignment module analysis}(\subref{fig: minimal or maximal optimization}). The experiment results show that the directional information further improves alignment accuracy. Moreover, adding the directional information makes minimal optimization becomes possible. Here, we found that the apply softmax operation when adding directional information is an unnecessary step and will decrease system accuracy. We adopt the softmax on matrix A and get 0.964, 0.976, and 0.997 alignment results in the Zh-En, Ja-En, and Fr-En datasets respectively. Compared with the results without softmax (0.966, 0.990, and 0.996). The softmax function does not bring benefit in Zh-En and Fr-En datasets and decreases 1.4\% accuracy in the Ja-En dataset. 

From the experiment results, we can conclude that both minimal and maximal optimization achieves outstanding alignment results. The minimum optimization has a strong reliance on the directional property and increases 61\%, 59\%, and 55\% accuracy in three datasets when adding directional property. In conclusion, our alignment method can adapt maximum or minimum value optimization and further improve alignment process flexibility.

\subsection{Additional Analysis}
\label{sec: additional analysis}
\begin{figure}
\centering
\begin{subfigure}[t]{0.495\linewidth}
    \includegraphics[width=\linewidth]{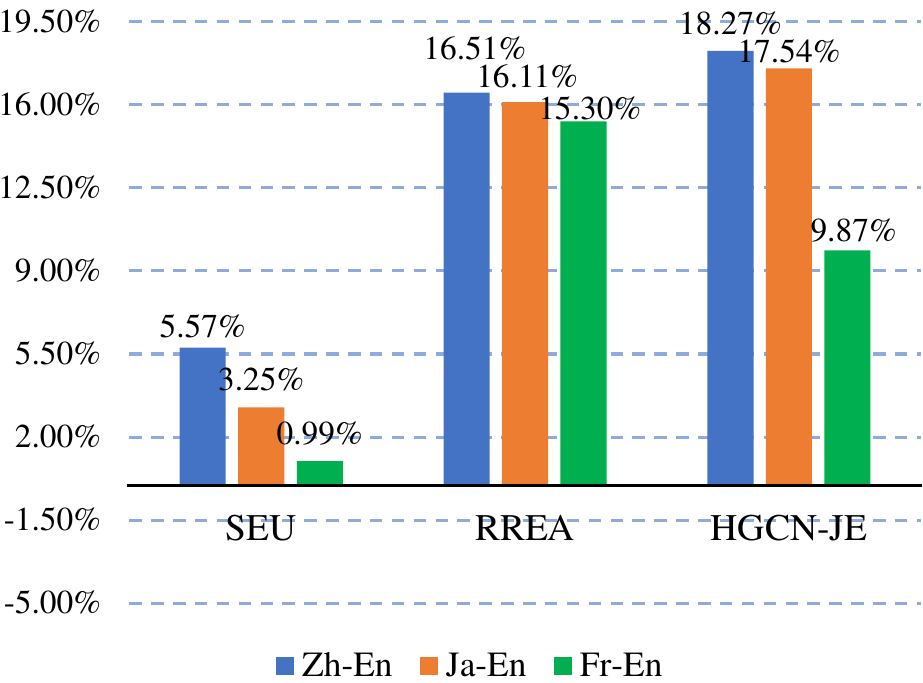}
    \subcaption{The performance gain of attaching our algorithm on improving existing entity alignment methods.} 
    \label{fig: improve existing alignment methods}
\end{subfigure}
\begin{subfigure}[t]{0.497\linewidth}
    \includegraphics[width=\linewidth]{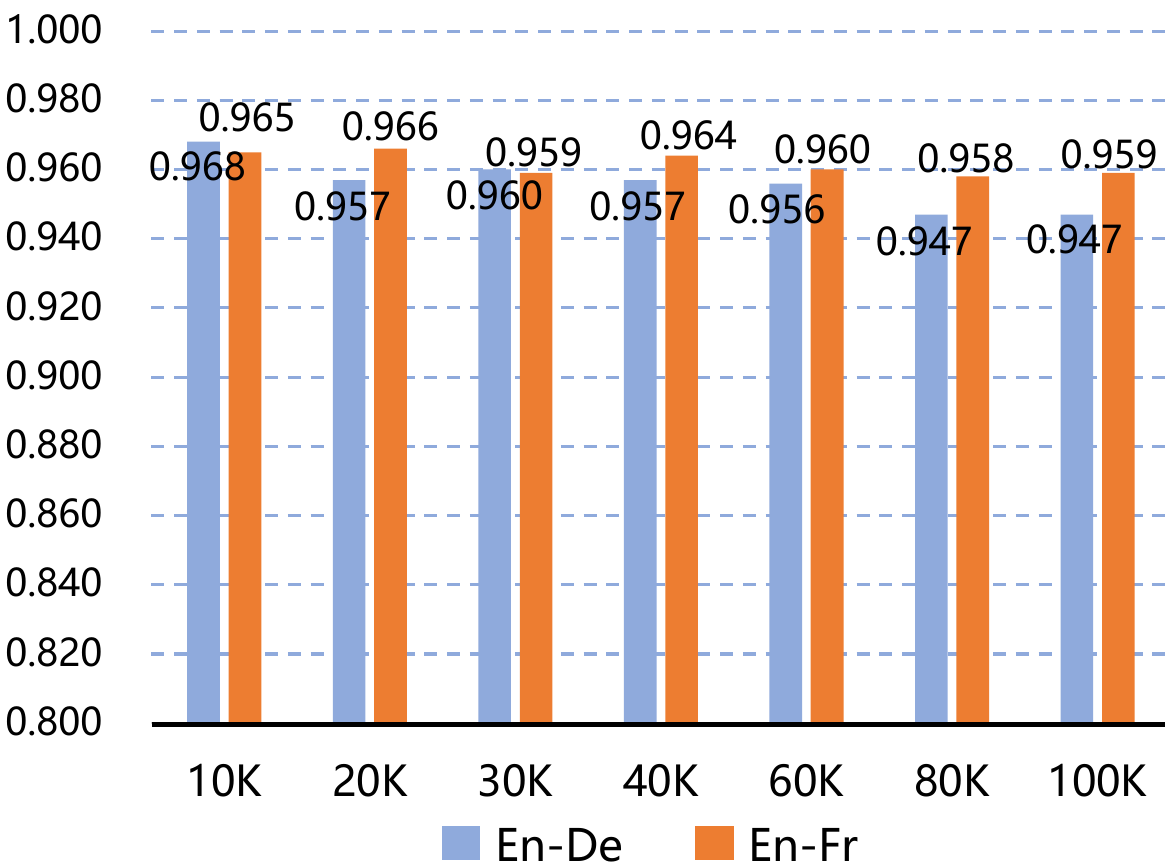}
    \subcaption{The adapting ability of our method on different data sizes.}
    \label{fig: facing data changes}
\end{subfigure}
\caption{The experiment results of additional analysis.}
\label{fig: additional analysis}
\end{figure}

We conduct two experiments to show the benefit our alignment strategy brings to the existing alignment methods. Three open source code methods SEU~\cite{seu}, RREA~\cite{rrea}, and HGCN-JE~\cite{hgcn-je-jr} are evaluated. To evaluate the adaptability of our alignment method to data size changes, we use a larger size dataset DBP100K~\cite{openea}. The DBP100K dataset includes En-De and En-Fr two cross-language pairs, and each language pair includes 100K aligned entity. We subsampled 10K, 20K, 30K, 40K, 60K, 80K, and 100K data to see the impact of different dataset sizes on our alignment method. The experiments are displayed in Fig.~\ref{fig: additional analysis}.

The results of Fig.~\ref{fig: additional analysis}(\subref{fig: improve existing alignment methods}) indicate that our alignment strategy brings improvement in all existing alignment methods. In those three methods, HGCN-JE received the highest received performance gain of 18.27\% accuracy in the Zh-En dataset. The evaluation for the adaptability is shown in Fig.~\ref{fig: additional analysis}(\subref{fig: facing data changes}). As the data size increases, so does the complexity of the data, the maximum performance drop of our method is less than 2.1\%.

\section{Error Mode Analysis}
\begin{CJK}{UTF8}{gbsn}
In this section, we are analyzing the entity pairs that our algorithm falsely aligned. Through our analysis and statistics, we list the main four circumstances as follows. 1) Abbreviation: one entity is the other entity abbreviation in aligned entity pairs, such as "société nationale des chemins de fer français $\leftrightarrow$ sncf". In fact, it's difficult to recognize even by an expert. 2) Person's name: In different nations, one person may have a different name. Besides, the same name also can correspond to a different person. Such as "李正秀 $\leftrightarrow$ lee jung-soo". 3) Drama show/Music record names: one drama show/music record may have a different official name in a different nation or region, which is due to language, culture, marketing et al. Besides that, records may have many informal names such as "静寂の世界 $\leftrightarrow$ a rush of blood to the head". 4) Special characters: some special characters may be due to system error or typo such as "『z』の誓い $\leftrightarrow$ \%22z\%22 no chikai". 
\end{CJK}

\section{Conclusion and Future Research}
In this paper, we propose an unsupervised entity alignment method. We first utilize the multi-language encoder combined with Google translator to encode the multi-view information which reduces the reliance on labeled data. Then we adopt the optimization-based method to align the entities. Our method can generate ranked matching results which provides a higher degree of flexibility in the matching process. Our method outperformed state of art methods in unsupervised and semi-supervised categories. Moreover, our novel alignment strategy can be used as a plug-in to bring in additional performance gains to existing methods. 

Our future work will be in two directions. Firstly, we used the off-the-shelf encoders which, at today's standard, it is considered a small-size language model. Developing a large-scale language encoder may bring additional benefits to our algorithm. Secondly, we can deploy our algorithm into medical applications where the patient record can be used as a query to conduct a multi-lingual search in the global dataset.

\section{Acknowledgement}
This work is supported by Hainan Province Science and Technology Special Fund (No.ZDKJ2021042), A*STAR under its Artificial Intelligence in Medicine Transformation (AIMx) Program (Award H20C6a0032) and National Natural Science Foundation of China(No.62362023).

\section{Ethical Statement}
This work doesn't involve any unethical tasks, all the experiment was done in public datasets.

\bibliographystyle{splncs04}
\bibliography{bibliography}
\end{document}